\documentclass[letterpaper, 10 pt, conference]{ieeeconf}
\IEEEoverridecommandlockouts
\usepackage{amsmath,amssymb,amsfonts,bm}
\usepackage{algorithmic}
\usepackage{graphicx}
\usepackage{textcomp}
\usepackage{xcolor}
\usepackage{lipsum}
\usepackage[style=ieee,dashed=false,isbn=false,url=false,doi=false]{biblatex}

\DeclareSourcemap{
  \maps{
    \map{
      \pertype{article}
      \step[fieldset=language, null]
      \step[fieldset=url, null]
      \step[fieldset=doi, null]
      \step[fieldset=issn, null]
      \step[fieldset=isbn, null]
      \step[fieldset=note, null]
      \step[fieldset=editor, null]
      \step[fieldset=urldate, null]
      \step[fieldset=file, null]
    }
  }
}
\DeclareSourcemap{
  \maps{
    \map{
      \pertype{inproceedings}
      \step[fieldset=language, null]
      \step[fieldset=url, null]
      \step[fieldset=doi, null]
      \step[fieldset=issn, null]
      \step[fieldset=isbn, null]
      \step[fieldset=note, null]
      \step[fieldset=editor, null]
      \step[fieldset=urldate, null]
      \step[fieldset=file, null]
    }
  }
}
\DeclareSourcemap{
  \maps{
    \map{
      \pertype{incollection}
      \step[fieldset=language, null]
      \step[fieldset=url, null]
      \step[fieldset=doi, null]
      \step[fieldset=issn, null]
      \step[fieldset=isbn, null]
      \step[fieldset=note, null]
      \step[fieldset=editor, null]
      \step[fieldset=urldate, null]
      \step[fieldset=file, null]
    }
  }
}

\def\BibTeX{{\rm B\kern-.05em{\sc i\kern-.025em b}\kern-.08em
    T\kern-.1667em\lower.7ex\hbox{E}\kern-.125emX}}

\begin{document}

\title{NMPC-based Unified Posture Manipulation\\ and Thrust Vectoring for Fault Recovery}

\author{
Adarsh Salagame$^{1\text{\textdagger}}$, Shashwat Pandya$^{1\text{\textdagger}}$, Ioannis Mandralis$^{2}$,\\Eric Sihite$^{2}$, Alireza Ramezani$^{1*}$, and Morteza Gharib$^{2}$%
\thanks{$^{1}$ The author is with the SiliconSynapse Laboratory, Department of Electrical and Computer Engineering, Northeastern University, Boston, MA-02119, USA. (e-mail: salagame.a, pandya.shas, a.ramezani@northeastern.edu).}%
\thanks{$^{2}$ The author is with the Department of Aerospace, California Institute of Technology, Pasadena, CA-91125, USA. (e-mail: esihite, imandralis, mgharib@caltech.edu).}%
\thanks{{\text{\textdagger}} These authors have equal contributions to this work
}
\thanks{{*} Corresponding author's e-mail: a.ramezani@northeastern.edu.} 
}
\maketitle

\begin{abstract}
Multi-rotors face significant risks, as actuator failures at high altitudes can easily result in a crash and the robot's destruction. Therefore, rapid fault recovery in the event of an actuator failure is necessary for the fault-tolerant and safe operation of unmanned aerial robots. In this work, we present a fault recovery approach based on the unification of posture manipulation and thrust vectoring. The key contributions of this work are: 1) Derivation of two flight dynamics models (high-fidelity and reduced-order) that capture posture control and thrust vectoring. 2) Design of a controller based on Nonlinear Model Predictive Control (NMPC) and demonstration of fault recovery in simulation using a high-fidelity model of the Multi-Modal Mobility Morphobot (M4) in Simscape.
\end{abstract}

\section{Introduction}

In aerial vehicles, any actuator or sensor failure can cause fatal problems and significantly increase the risk of crashing. Transport aircraft implement redundancy in their propulsion systems, sensors, controls, and even human operators to minimize the risk of crashing, stalling, and uncontrollable states, which can save lives and avoid costly damages.

Multi-rotors face similar risks, as actuator failures at high altitudes can easily result in a crash and the robot's destruction. Therefore, rapid fault detection and a robust controller that can stabilize the robot in the event of an actuator failure are necessary for the fault-tolerant, safe operation of unmanned aerial robots.

Much research has been conducted on fault detection and fault-tolerant control methods in quadrotors. Fault detection methods leverage vibrational analysis \cite{ghalamchi2018vibration}, sensor fusion \cite{pourpanah2018anomaly}, and performance monitoring \cite{palanisamy2022fault} to detect propeller anomalies. Vibration-based approaches can identify imbalances, while condition monitoring enhances early failure detection.

For fault-tolerant control, nonlinear model predictive control (MPC) \cite{nan2022nonlinear} and adaptive learning-based methods \cite{oconnell2024learning} have been shown to improve stability under actuator faults. Quadrotors can maintain controllability despite losing multiple propellers by using differential thrust \cite{mueller2014stability}, adaptive vision-based strategies \cite{sun2021autonomous}, and model-based estimation \cite{mao2024propeller}. Reinforcement learning enhances resilience by learning optimal control policies under failures \cite{liu2024reinforcement}. Passive fault-tolerant techniques \cite{ke2023uniform} provide robustness without explicit fault diagnosis. Tilting-rotor quadcopters offer an alternative fault-tolerant mechanism by leveraging thrust vectoring to maintain stability under propeller failures \cite{nemati2016stability}. Control design methods such as adaptive actuator fault mitigation \cite{mallavalli2019fault} and robust detection with fixed-time recovery \cite{mazare2024robust} enhance UAV reliability. Neural adaptive controllers \cite{abbaspour2020neural} further enable active fault compensation, while multirotor transportation systems can operate with blade damage using resilient control strategies \cite{yu2024fault}. High-accuracy adaptive robust controllers address actuator uncertainties and aerodynamic drag \cite{liang2024high}. Additionally, formation control of multi-UAVs under actuator faults can be achieved with fixed-time collision-free approaches \cite{miao2024fixed}. Event-triggered adaptive fuzzy controllers provide fault tolerance under actuator saturation \cite{wang2024event}. Tilt tri-rotor UAVs benefit from fault-tolerant position tracking methods \cite{hao2022fault}, while hybrid reinforcement learning techniques improve robustness in UAV control \cite{sohege2021novel}.

\begin{figure}[t]
    \centering
    \vspace{0.1in}
    \includegraphics[width=\linewidth]{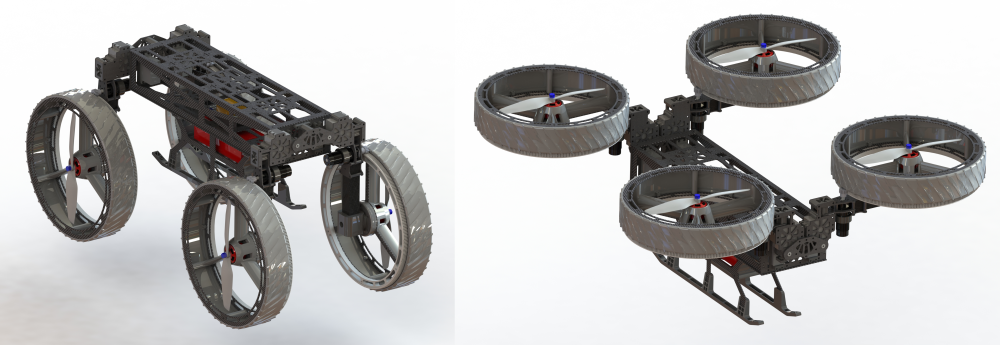}
    \caption{Multi-Modal Mobility Morphobot (M4), a versatile transforming robot capable of changing modes into rover (UGV) mode for wheeled ground locomotion, quadrotor (UAS) mode for aerial locomotion, and other modes depending on environmental challenges and energy efficiency demands.}
    \vspace{-0.1in}
    \label{fig:cover}
\end{figure}

In this work, we aim to implement a fault-tolerant flight controller from a different perspective: the combination of posture manipulation and thrust vectoring. In this approach, the control matrix that maps thruster wrench to the states depends on the time-varying shape of the vehicle.

The model we use to test this idea is the Multi-Modal Mobility Morphobot (M4) \cite{sihite_multi-modal_2023, mandralis2023minimum, sihite2024dynamic,}, shown in Fig.~\ref{fig:cover}. M4 possesses extensive locomotion plasticity; that is, the robot can achieve distinct modes of mobility, including wheeling, aerial maneuvering, and legged mobility. The underlying design concept that yields this plasticity is the combination of function (leg, wheel, thruster) and a morphing structure (a design with 8 body joints). Therefore, the mobility of M4 can take combinatorial forms that encompass all combinations of functions and postures.

The overarching objective of this paper is to leverage M4's ability to integrate function and posture control (e.g., thruster-assisted slope locomotion inspired by birds \cite{sihite2024posture,pitroda_enhanced_2024,pitroda_quadratic_2024,krishnamurthy_enabling_2024,salagame2024quadrupedal,de_oliveira_thruster-assisted_2020,sihite_optimization-free_2021,krishnamurthy2024narrow,dangol_performance_2020,dangol_control_2021}) to achieve fault-tolerant flight. The contributions of this work are:

The contributions of this work are:

\begin{itemize}
    \item Proposal of a new fault recovery approach based on the combined use of posture manipulation and thrust vectoring.
    \item Derivation of two flight dynamics models (high-fidelity and reduced-order) that capture posture control and thrust vectoring.
    \item Design of a controller based on Nonlinear Model Predictive Control (NMPC) and demonstration of fault recovery in simulation using a high-fidelity model of M4 in Simscape.
\end{itemize}

This work is organized as follows. First, brief overview of M4 is provided. Then, we describe the modeling approach followed by control design based on NMPC. Last, we present the results, discussion, and concluding remarks.

\section{Quick Design Overview of M4}

M4's body possesses 18 degrees of freedom (DoF). In its design, each leg has 3 DoFs. Two joints attach the legs to the body and allow two types of movement: 1) swing and 2) abduction-adduction (sideways) movement. A wheel is affixed to the end of each leg. The wheels feature 3D-printed gear meshes and are driven by servo actuators on each leg.

Each wheel, at its hub, is equipped with a brushless DC motor and a propeller (a thruster). Therefore, the appendages in M4 can change function from leg (engaging the gear mesh) to wheel (fixing the legs and actuating the gear mesh servo) or to thruster (activating the brushless DC motor). The choice of locomotion (e.g., ground vs. aerial) depends on the objectives, as this robot can opt to minimize energy consumption using ground locomotion, or employ the power-hungry aerial mode to bypass obstacles that cannot be easily navigated on the ground.

M4 weighs approximately 6.0 kg, including all components: onboard computers for low-level control and data collection, sensors (encoders, inertial measurement unit, stereo cameras), communication devices for teleoperation, joint actuators, propulsion motors, power electronics, and battery. In rover mode, M4 measures 0.7 m in length and 0.35 m in both width and height, while in quadrotor mode the propeller axis distance is 0.45 m. The propellers can generate a maximum thrust of 8.8 kg, resulting in an approximate thrust-to-weight ratio of 1.5.

\section{Flight Dynamics Model}

We model the high-fidelity system and reduced-order model which will serve as the prediction model 

\begin{figure}[t]
    \centering
    \vspace{0.1in}
    \includegraphics[width=0.8\linewidth]{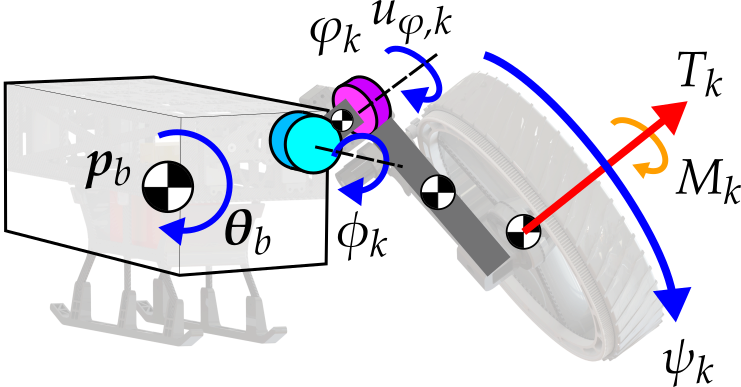}
    \caption{Free-body Diagram of M4, showing the main body and one leg. Each leg has two hip joints (frontal and sagittal), wheel joint which is driven by a motor, and a thruster that generates thrust force and moment for flight.}
    \vspace{-0.1in}
    \label{fig:fbd}
\end{figure}

We consider the general state-space form for the flight dynamics of our vehicle:
\begin{equation}
\dot{x} = f(x,u),    
\end{equation}
\noindent where \( x \) is the state vector and \( u \) is the control input. For the high-fidelity model, the system is modeled as a collection of rigid bodies representing the main body and its appendages (three lump of mass considered per each leg). Define the generalized coordinates as
\begin{equation}
q = \begin{bmatrix} p_b, \theta_b, q_a \end{bmatrix}^\top,
\end{equation}
\noindent where \( p_b \in \mathbb{R}^3 \) is the position of the main body, \( \theta_b \in \mathbb{R}^3 \) represents its orientation (e.g., Euler angles), and \( q_a \) collects the joint angles of the appendages. The state vector is given by
\begin{equation}
x = \begin{bmatrix} q,\dot{q} \end{bmatrix}^\top.
\end{equation}

The dynamics are derived using the Euler--Lagrange formulation. Let the Lagrangian be
\begin{equation}
L(q,\dot{q}) = K(q,\dot{q}) - V(q),
\end{equation}
where \( K \) is the total kinetic energy (of both the main body and the appendages) and \( V \) is the gravitational potential energy. The Euler--Lagrange equation is
\begin{equation}
\frac{d}{dt}\left(\frac{\partial L}{\partial \dot{q}}\right) - \frac{\partial L}{\partial q} = Q,
\end{equation}
with \( Q \) representing the generalized forces, which include contributions from joint torques (\(\tau_j\)), thruster forces (\(T_k\)), and reactive moments (\(M_k\)). These inputs are related to the generalized coordinates by a mapping matrix \( B(q) \), so that
\begin{equation}
Q = B(q)\, u, \quad \text{with} \quad u = \begin{bmatrix} T_k,\tau_j\end{bmatrix}^\top.
\end{equation}
The resulting equations of motion can be written in the standard second-order form as
\begin{equation}
M(q)\ddot{q} + C(q,\dot{q})\dot{q} + g(q) = B(q)\, u,
\end{equation}
where \( M(q) \) is the mass/inertia matrix, \( C(q,\dot{q}) \) represents Coriolis and centrifugal terms, and \( g(q) \) collects the gravitational forces. Defining the state as
\begin{equation}
x = \begin{bmatrix} q, \dot{q} \end{bmatrix}^\top,
\end{equation}
the state-space representation is
\begin{equation}
\dot{x} = \begin{bmatrix}
\dot{q} \\
M(q)^{-1}\Big( B(q)\, u - C(q,\dot{q})\,\dot{q} - g(q) \Big)
\end{bmatrix}.
\end{equation}

\subsection{Prediction Model}

The prediction model used for control design is obtain as follows. The main body \((m_b)\) is treated as a 6-DoF rigid body and the appendages are assumed to have point mass \((m_l)\), influencing the mapping of thruster forces. The state vector for the ROM is defined as
\begin{equation}
x_{\text{rom}} = \begin{bmatrix} p_b,\theta_b, q_a, \dot{p}_b,\omega_b, \dot{q}_a \end{bmatrix}^\top,
\end{equation}
where \( p_b \in \mathbb{R}^3 \) is the position of the main body, \( \theta_b \in \mathbb{R}^3 \) represents its orientation (e.g., via Euler angles), \( \dot{p}_b \in \mathbb{R}^3 \) is the linear velocity, and \( \omega_b \in \mathbb{R}^3 \) is the angular velocity.

The translational dynamics are given by Newton's second law:
\begin{equation}
m_{net} \ddot{p}_b = f_{\text{ext}}(q_a,u),
\end{equation}
where \( m_{net} \) \((m_b +4m_l\)) is the total mass of the system and \( f_{\text{ext}}(q_a,u) \) includes the net thruster force (adjusted by the appendage configuration \( q_a \)) and gravity. The rotational dynamics follow the Newton--Euler formulation:
\begin{equation}
I_b \dot{\omega}_b + \omega_b \times (I_b \omega_b) = \tau_{\text{ext}}(q_a,u),
\end{equation}
with \( I_b \) the inertia matrix of the main body and \( \tau_{\text{ext}}(q_a,u) \) the net external moment (again, influenced by \( q_a \) and the thruster inputs). The kinematic relationship between the Euler angle rates and the angular velocity is expressed as
\begin{equation}
\dot{\theta}_b = J(\theta_b)\,\omega_b,
\end{equation}
where \( J(\theta_b) \) is the transformation matrix determined by the Euler angle convention.

Thus, the state-space equations for the ROM are:
\[
\dot{p}_b = \dot{p}_b, \qquad
\dot{\theta}_b = J(\theta_b)\,\omega_b,
\]
\[
\ddot{p}_b = \frac{1}{m_{net}}\Big( f_{\text{thrusters}}(q_a,u) + m_b g \Big),
\]
\[
\dot{\omega}_b = I_b^{-1}\Big( \tau_{\text{thrusters}}(q_a,u) - \omega_b \times (I_b \omega_b) \Big).
\]
In compact form, the ROM state-space model is:
\begin{equation}
\dot{x}_{\text{rom}} = \begin{bmatrix}
\dot{p}_b \\
J(\theta_b)\,\omega_b \\
\dot q_a \\
\frac{1}{m_b}\Big( f_{\text{thrusters}}(q_a,u) + m_b g \Big) \\
I_b^{-1}\Big( \tau_{\text{thrusters}}(q_a,u) - \omega_b \times (I_b \omega_b) \Big) \\
J_a \, u
\end{bmatrix},
\end{equation}
where $J_a$ maps $u$ into the joint acceleration.

\section{Fault Recovery by Nonlinear Model Predictive Control (NMPC)}

The fault recovery controller uses the thrusters and hip sagittal joint torques to stabilize the robot's orientation in case one of the thrusters fails. Only the hip sagittal actuators are used to maintain a planar flight configuration where all thrusters point upwards in the robot's z-axis. Therefore, we can simplify the equations of motion by reducing the inputs to only include the actuators used in the fault recovery process. 

We formulate the NMPC problem as follows:
\begin{equation}
\begin{aligned}
\min_{\{u_j\}_{j=0}^{N_h-1}} \quad & J = \sum_{j=0}^{N_h-1} \Big[ \big(x_j - x_{\text{ref},j}\big)^T Q \big(x_j - x_{\text{ref},j}\big) \\
& \qquad \qquad \qquad+ u_j^T R u_j \Big] \\
\text{subject to} \quad & x_{j+1} = \Phi(x_j,u_j), \quad j=0,\ldots,N_h-1, \\
& x_0 = x(t_0), \\
& x_{\min} \le x_j \le x_{\max}, \quad u_{\min} \le u_j \le u_{\max},
\end{aligned}
\end{equation}
where \( N_h \) is the prediction horizon and the function \( \Phi(\cdot,\cdot) \) represents the discrete-time evolution of the ROM dynamics over one time step, which may be obtained via a numerical integration scheme (e.g., fourth-order Runge--Kutta) applied to
\begin{equation}
\dot{x}_{\text{rom}} = f_{\text{rom}}(x_{\text{rom}},u).
\label{eq:mpc_model}    
\end{equation}
The matrices \( Q \) and \( R \) are weighting matrices penalizing the state tracking error and control effort, respectively, while \( x_{\text{ref},j} \) denotes the desired reference trajectory at step \( j \).

\section{Simulation Results}

\begin{figure}[t]
    \centering
    \vspace{0.1in}
    \includegraphics[width=1\linewidth]{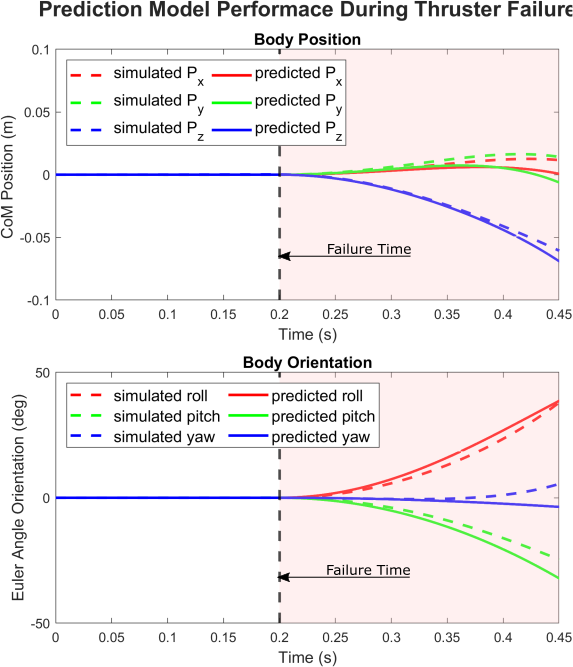}
    \caption{Plot showing simulated states between the NMPC prediction model states and Simscape model during thruster failure event. The predicted model closely resembles the full fidelity model.}
    \vspace{-0.1in}
    \label{fig:plot_prediction}
\end{figure}

\begin{figure}[t]
    \centering
    \vspace{0.1in}
    \includegraphics[width=1\linewidth]{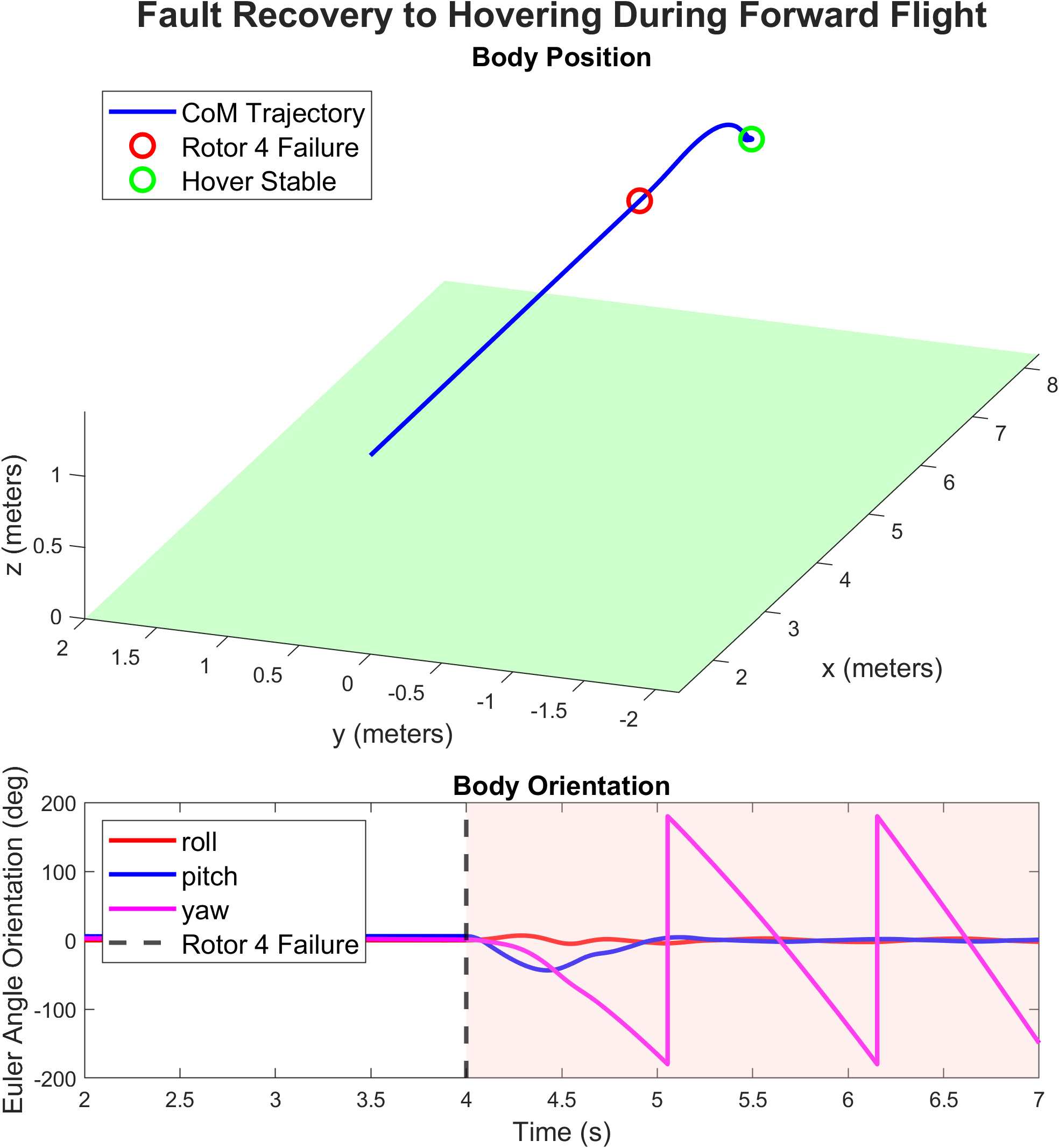}
    \caption{Plots showing the center of mass trajectory and Euler angles of the robot during the failure event until it reached a stable recovery state. The failure timing is shown with the red circle and dashed line.}
    \vspace{-0.1in}
    \label{fig:plot_states_1}
\end{figure}

\begin{figure}[t]
    \centering
    \vspace{0.1in}
    \includegraphics[width=1\linewidth]{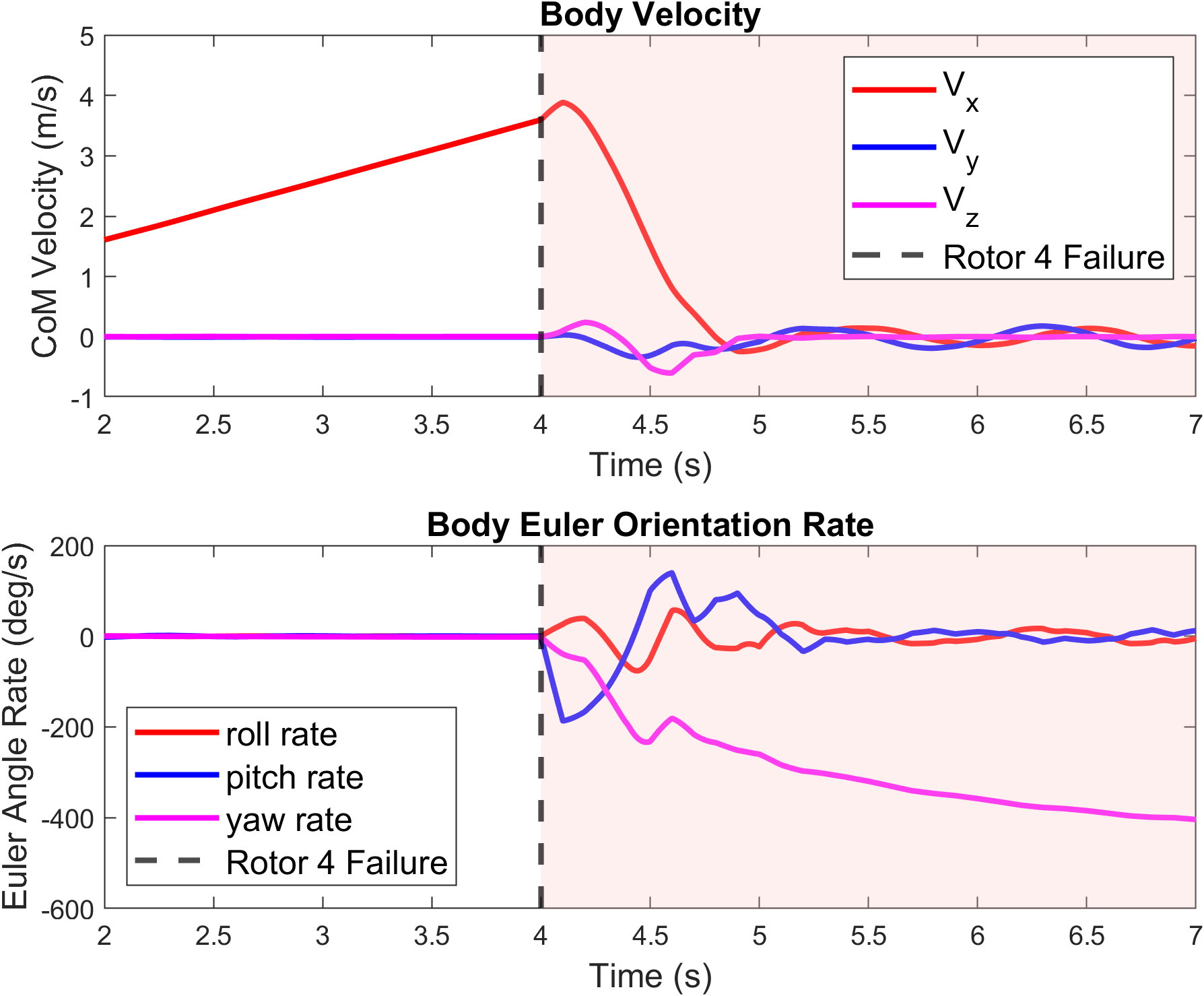}
    \caption{Plot showing the simulated body linear and angular velocities from the beginning with fully functional thrusters, thruster failure event, and recovery states. The yaw is unstable due to the loss of a thuster and the NMPC prioritizing roll and yaw stability during the failure recovery phase.}
    \vspace{-0.1in}
    \label{fig:plot_velocity_1}
\end{figure}

\begin{figure}[t]
    \centering
    \vspace{0.1in}
    \includegraphics[width=1\linewidth]{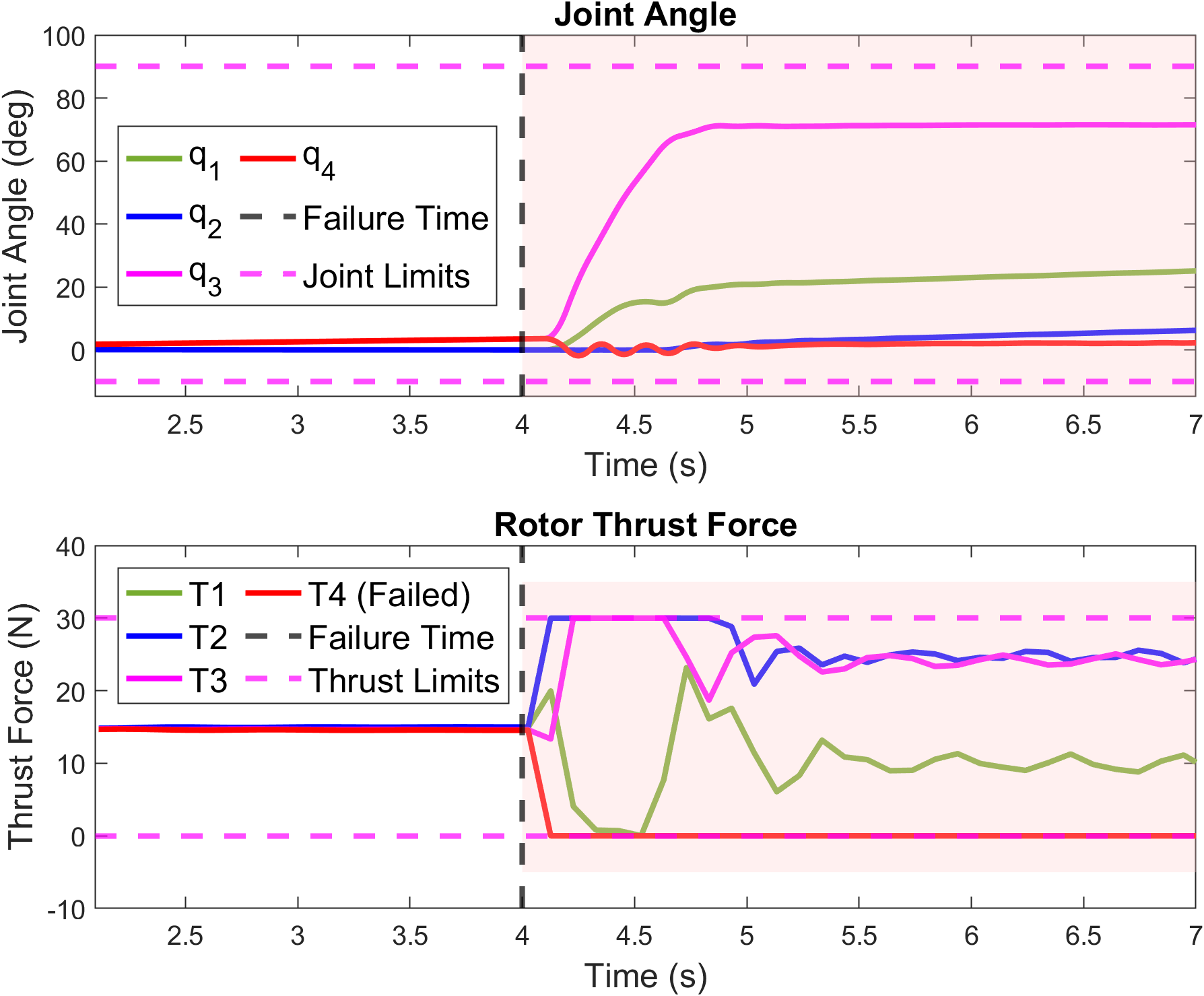}
    \caption{Plots showing the control inputs (joint angle and thruster values) in the simulation as calculated by the NMPC during the thruster failure event. Thruster 4 force goes to zero during the failure event.}
    \vspace{-0.1in}
    \label{fig:plot_control_1}
\end{figure}

\begin{figure*}[t]
    \centering
    \vspace{0.1in}
    \includegraphics[width=1\linewidth]{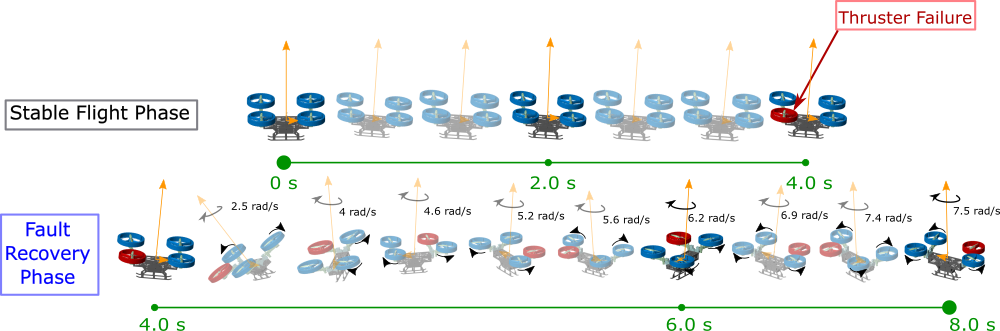}
    \caption{Illustrations of the robot during the flight, showing a stable flight without and with thruster failure (shown in red). The fault recovery phase and trajectory are shown in the bottom row.}
    \vspace{-0.1in}
    \label{fig:render}
\end{figure*}

\begin{figure}[t]
    \centering
    \vspace{0.1in}
    \includegraphics[width=1\linewidth]{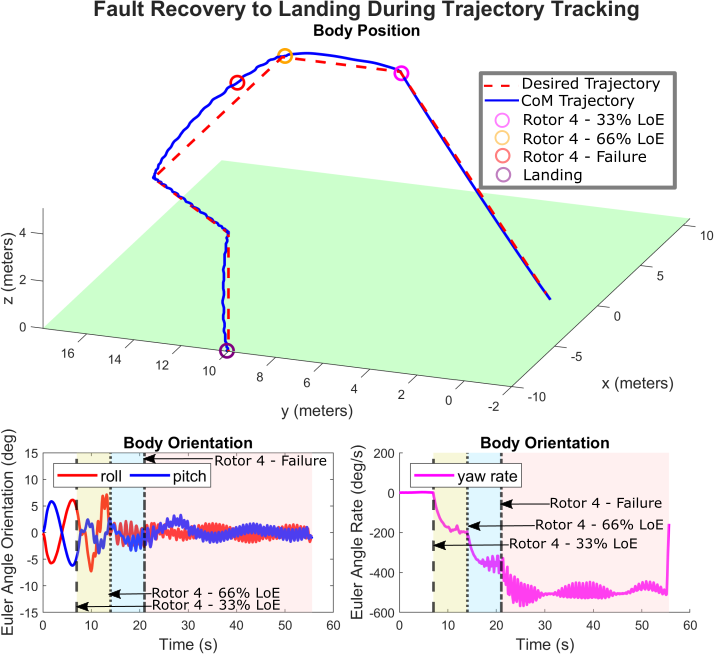}
    \caption{Plots showing the center of mass trajectory and Euler angles of the robot during the multiple failure events. The robot attempts to track the reference trajectory despite the failure and successfully lands.}
    \vspace{-0.1in}
    \label{fig:plot-tracking}
\end{figure}

\begin{figure}[t]
    \centering
    \vspace{0.1in}
    \includegraphics[width=1\linewidth]{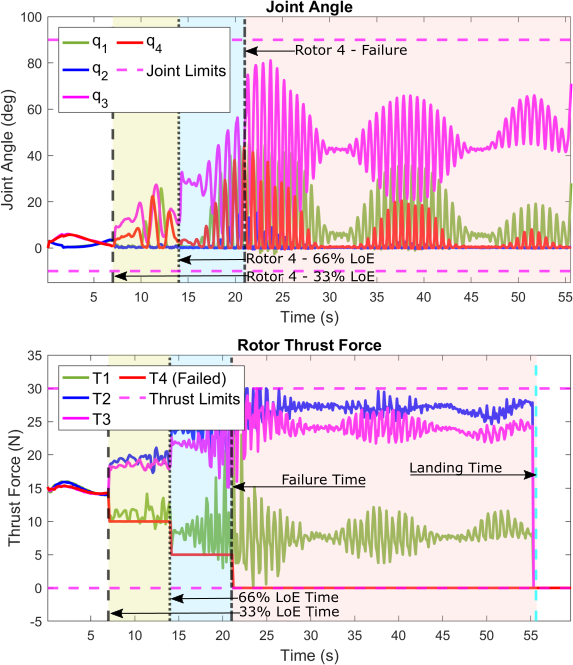}
    \caption{Plot for thrust forces and joint angles for trajectory tracking during multiple loss of power events of varying severity. Rotor 4 undergoes 33\% LoE from 7 s to 14 s, 66\%LoE from 14 s to 21 s, and complete failure after 21 s. The plots show the evolution of the controller values through multiple degrees of thruster failures.}
    \vspace{-0.1in}
    \label{fig:plot-input}
\end{figure}





The simulation was performed in Simscape, a rigid body simulator within MATLAB's Simulink environment. By inputting the controller values into the model, the simulator computes the robot's states, which are then fed into the NMPC optimizer along with the reference states. The NMPC uses a constant cost weighting matrix and fixed parameters to determine the optimal controller input values, which are applied in the Simscape model to advance the state forward in time. The thruster moment is assumed to be proportional to the thrust force of the same rotor, as well as implementing the aerodynamic drag to the robot's body. This enables yaw rate to be more realistic and reach a saturation compared to a model without any drag.

The NMPC's prediction model utilizes the reduced-order model of \eqref{eq:mpc_model} to minimize computational cost. The state is propagated forward in time within the NMPC using a 4th-order Runge–Kutta scheme while enforcing the state and input constraints. 


The NMPC prediction model runs with a fixed time step of 0.1 s and a prediction horizon of 5 steps. The controller input determined by the NMPC remains constant within each time step and is updated every 0.1 s, matching the NMPC prediction model's time step. 

The primary state constraints in the NMPC are applied to body orientation and joint angles. Specifically, roll and pitch are constrained within \([-90, 90]\) degrees, while the hip sagittal joint is limited to \([0, 90]\) degrees, where 90 degrees corresponds to the leg being fully extended outward. 

Additionally, an extra constraint is imposed on the joint angles to prevent wheel collisions, ensuring that the sum of joint angles on each side of the body remains below approximately 110 degrees to avoid interference. The thrust forces are constrained to \([0, 30]\) N, and joint accelerations are limited to \([-50, 50]\) rad/s$^2$. No explicit constraint is imposed on yaw, allowing the robot to rotate freely about the yaw axis in the event of thruster failure, while the remaining states remain unconstrained.


The controller has no prior knowledge of faults, eliminating the need for controller switching. The proposed nonlinear MPC controller functions both as a stable flight controller and as a fault-tolerance controller. To validate this, two simulations were performed in different phases. In these simulations, the robot is initially in a stable, fault-free flight phase. Then, the maximum thrust of rotor 4 is reduced to simulate a thruster failure scenario. We assume that, in a real-world scenario, the controller would detect this thrust reduction based on sensor information and the inability to exceed or saturate thrust values. 

In the first scenario, the robot transitions from stable flight to fault recovery. In the second scenario, the robot follows a predefined trajectory, gradually losing thrust until complete failure, after which it proceeds to land safely.


The simulation result presented in Fig.~\ref{fig:plot_prediction} shows the performance of the prediction model during a thruster failure. The robot is in stable hovering condition for the initial 0.2 s. Rotor 4 is allowed to switch off at 0.2 s and the position and Euler angles after this instant are captured in the presented figure. The comparison is made between the reduced order prediction model and the full fidelity Simscape model. It can be seen that the dynamic behavior of the prediction model closely resembles the dynamics of the full fidelity Simscape model after the instance of the thruster failure. 

Thereafter, we simulate the scenario when the robot is given an initial trajectory in the x-direction, thus performing forward flight to traverse from the initial position to a goal position, with the condition that in the event of thruster failure, it should maintain its current position. Thruster 4 fails at time instance of 4 s and the results of body position, Euler angles, body velocities, and Euler angle rates are presented in Fig.~\ref{fig:plot_states_1} and Fig.~\ref{fig:plot_velocity_1}. The robot is traversing at over 3 m/s right before the thruster failure and thereafter, it starts to rotate about its primary axis, and enters a stable configuration with increasing yaw rate within 1 s. The thrust forces and joint angles for this scenario are presented in Fig.~\ref{fig:plot_control_1}. This demonstrates that robot is capable of successfully recovering from a thruster failure at any instant. For better visualization, the sequential snapshots of stable phase and fault recovery phase for this scenario are presented in Fig.~\ref{fig:render}. The snapshots in the first row of Fig.~\ref{fig:render} corresponds to stable flight phase where the robot pitches forward to traverse for 4 s. Thereafter at 4 s, failure for rotor 4 takes place, after which the robot undergoes instantaneous changes in roll, pitch and yaw. By using sagittal joints and thrust vectoring, it stabilizes roll and pitch rapidly while achieving a yaw rate at the end of 4 s.

Lastly, we simulate the scenario of trajectory tracking with multiple way points. The trajectory tracking is performed under multiple phases where the robot is initially in the stable flight phase, followed by 33\% Loss of Effectiveness (LoE), 66\% LoE, complete failure of rotor 4, and landing phase. LoE, in this scenario, is defined as percentage loss of thrust relative to hover thrust. Also, the robot was allowed to stabilize at each way point for about 1 s before proceeding to the next way point of the trajectory. The plots for position tracking as well as body orientation are presented in Fig.~\ref{fig:plot-tracking}. As soon as the LoE occurs, the robot starts spinning about its primary axis, with gradually increasing yaw rate. After complete failure, the yaw saturation occurs within 10 s. It can be seen from the position tracking plot in Fig.~\ref{fig:plot-tracking}, that the robot closely follows the reference trajectory even after partial or complete failure of rotor 4. Finally, in the last phase, the robot descends along z to successfully land on the ground, as thrust forces are shut off and yaw rate starts to converge rapidly. Figure~\ref{fig:plot-input} shows the thrust forces and changes in sagittal joint angles throughout the entire trajectory tracking simulation.





\section{Conclusions}

In this work, we studied fault recovery in the flight mode of M4 by leveraging the robot's morphing characteristics. We achieved accurate model matching between Simscape and the prediction model after the instance of thruster failure. We demonstrated the ability to rapidly recover and maintain altitude after a thruster failure at cruising speed, enabling fault tolerance without controller switching. Additionally, we showed that even after failure, highly accurate trajectory tracking is possible, ensuring a safe landing.  

We also demonstrated that by utilizing posture manipulation, it is possible to significantly increase the time taken to reach yaw saturation. Future research will focus on hardware implementation and validating the results through real-world experiments.

\printbibliography

@misc{pitroda_enhanced_2024,
    title = {Enhanced {Capture} {Point} {Control} {Using} {Thruster} {Dynamics} and {QP}-{Based} {Optimization} for {Harpy}},
    url = {http://arxiv.org/abs/2411.17727},
    doi = {10.48550/arXiv.2411.17727},
    urldate = {2025-03-12},
    publisher = {arXiv},
    author = {Pitroda, Shreyansh and Sihite, Eric and Liu, Taoran and Krishnamurthy, Kaushik Venkatesh and Wang, Chenghao and Salagame, Adarsh and Nemovi, Reza and Ramezani, Alireza and Gharib, Morteza},
    month = nov,
    year = {2024},
    note = {arXiv:2411.17727 [cs]},
}

@misc{pitroda_quadratic_2024,
    title = {Quadratic {Programming} {Optimization} for {Bio}-{Inspired} {Thruster}-{Assisted} {Bipedal} {Locomotion} on {Inclined} {Slopes}},
    url = {http://arxiv.org/abs/2411.12968},
    doi = {10.48550/arXiv.2411.12968},
    urldate = {2025-03-12},
    publisher = {arXiv},
    author = {Pitroda, Shreyansh and Sihite, Eric and Krishnamurthy, Kaushik Venkatesh and Wang, Chenghao and Salagame, Adarsh and Nemovi, Reza and Ramezani, Alireza and Gharib, Morteza},
    month = nov,
    year = {2024},
    note = {arXiv:2411.12968 [cs]},
}

@misc{krishnamurthy_enabling_2024,
    title = {Enabling steep slope walking on {Husky} using reduced order modeling and quadratic programming},
    url = {http://arxiv.org/abs/2411.11788},
    doi = {10.48550/arXiv.2411.11788},
    urldate = {2025-03-12},
    publisher = {arXiv},
    author = {Krishnamurthy, Kaushik Venkatesh and Sihite, Eric and Wang, Chenghao and Pitroda, Shreyansh and Salagame, Adarsh and Ramezani, Alireza and Gharib, Morteza},
    month = nov,
    year = {2024},
    note = {arXiv:2411.11788 [cs]},
}

@inproceedings{de_oliveira_thruster-assisted_2020,
	title = {Thruster-assisted {Center} {Manifold} {Shaping} in {Bipedal} {Legged} {Locomotion}},
	doi = {10.1109/AIM43001.2020.9158967},
	booktitle = {2020 {IEEE}/{ASME} {International} {Conference} on {Advanced} {Intelligent} {Mechatronics} ({AIM})},
	author = {de Oliveira, Arthur C. B. and Ramezani, Alireza},
	month = jul,
	year = {2020},
	note = {ISSN: 2159-6255},
	pages = {508--513},
}

@article{sihite_multi-modal_2023,
	title = {Multi-{Modal} {Mobility} {Morphobot} ({M4}) with appendage repurposing for locomotion plasticity enhancement},
	volume = {14},
	copyright = {2023 Springer Nature Limited},
	issn = {2041-1723},
	url = {https://www.nature.com/articles/s41467-023-39018-y},
	doi = {10.1038/s41467-023-39018-y},
	language = {en},
	number = {1},
	urldate = {2023-10-07},
	journal = {Nature Communications},
	author = {Sihite, Eric and Kalantari, Arash and Nemovi, Reza and Ramezani, Alireza and Gharib, Morteza},
	month = jun,
	year = {2023},
	note = {Number: 1
Publisher: Nature Publishing Group},
	pages = {3323},
}

@inproceedings{dangol_performance_2020,
	title = {Performance satisfaction in {Midget}, a thruster-assisted bipedal robot},
	doi = {10.23919/ACC45564.2020.9147448},
	booktitle = {2020 {American} {Control} {Conference} ({ACC})},
	author = {Dangol, Pravin and Ramezani, Alireza and Jalili, Nader},
	month = jul,
	year = {2020},
	note = {ISSN: 2378-5861},
	pages = {3217--3223},
}

@inproceedings{sihite_optimization-free_2021,
	title = {Optimization-free {Ground} {Contact} {Force} {Constraint} {Satisfaction} in {Quadrupedal} {Locomotion}},
	doi = {10.1109/CDC45484.2021.9683155},
	booktitle = {2021 60th {IEEE} {Conference} on {Decision} and {Control} ({CDC})},
	author = {Sihite, Eric and Dangol, Pravin and Ramezani, Alireza},
	month = dec,
	year = {2021},
	note = {ISSN: 2576-2370},
	pages = {713--719},
}

@article{dangol_control_2021,
	title = {Control of {Thruster}-{Assisted}, {Bipedal} {Legged} {Locomotion} of the {Harpy} {Robot}},
	volume = {8},
	issn = {2296-9144},
	url = {https://www.frontiersin.org/articles/10.3389/frobt.2021.770514},
	urldate = {2023-05-17},
	journal = {Frontiers in Robotics and AI},
	author = {Dangol, Pravin and Sihite, Eric and Ramezani, Alireza},
	year = {2021},
}

@article{nan2022nonlinear,
  title={Nonlinear MPC for quadrotor fault-tolerant control},
  author={Nan, Fang and Sun, Sihao and Foehn, Philipp and Scaramuzza, Davide},
  journal={IEEE Robotics and Automation Letters},
  volume={7},
  number={2},
  pages={5047--5054},
  year={2022},
  publisher={IEEE}
}

@article{oconnell2024learning,
  title={Learning-Based Minimally-Sensed Fault-Tolerant  Adaptive Flight Control},
  author={O'Connell, Michael and Cho, Joshua and Anderson, Matthew and Chung, Soon-Jo},
  journal={IEEE Robotics and Automation Letters}, 
  title={Learning-Based Minimally-Sensed Fault-Tolerant Adaptive Flight Control}, 
  year={2024},
  volume={9},
  number={6},
  pages={5198-5205},
  doi={10.1109/LRA.2024.3389414}}

@inproceedings{mueller2014stability,
  author={Mueller, Mark W. and D'Andrea, Raffaello},
  booktitle={2014 IEEE International Conference on Robotics and Automation (ICRA)}, 
  title={Stability and control of a quadrocopter despite the complete loss of one, two, or three propellers}, 
  year={2014},
  volume={},
  number={},
  pages={45-52},
  doi={10.1109/ICRA.2014.6906588}}

@article{sun2021autonomous,
  title={Autonomous Quadrotor Flight Despite Rotor Failure With Onboard Vision Sensors: Frames vs. Events},
  author={Sihao Sun and Giovanni Cioffi and Coen de Visser and Davide Scaramuzza},
  journal={IEEE Robotics and Automation Letters},
  year={2021},
  volume={6},
  pages={580-587},
  url={https://api.semanticscholar.org/CorpusID:231682212}
}

@article{mao2024propeller,
author = {Jeffrey Mao and Jennifer Yeom and Suraj Nair and Giuseppe Loianno},
title = {From Propeller Damage Estimation and Adaptation to Fault Tolerant Control: Enhancing Quadrotor Resilience},
journal = {IEEE Robotics and Automation Letters},
year = {2024},
volume = {9},
publisher = {Institute of Electrical and Electronics Engineers (IEEE)},
month = {may},
url = {https://ieeexplore.ieee.org/document/10478191/},
number = {5},
pages = {4297--4304},
doi = {10.1109/lra.2024.3380923}
}

@article{liu2024reinforcement,
  title={Reinforcement Learning-Based Fault-Tolerant Control for Quadrotor UAVs Under Actuator Fault},
  author={Xiaoxu Liu and Zike Yuan and Zhiwei Gao and Wenwei Zhang},
  journal={IEEE Transactions on Industrial Informatics},
  year={2024},
  volume={20},
  pages={13926-13935},
  url={https://api.semanticscholar.org/CorpusID:272418617}
}

@article{ke2023uniform,
author = {Ke, Chenxu and Cai, Kai-Yuan and Quan, Quan},
title = {Uniform Passive Fault-Tolerant Control of a Quadcopter With One, Two, or Three Rotor Failure},
year = {2023},
issue_date = {Dec. 2023},
publisher = {IEEE Press},
volume = {39},
number = {6},
issn = {1552-3098},
url = {https://doi.org/10.1109/TRO.2023.3297048},
doi = {10.1109/TRO.2023.3297048},
journal = {Trans. Rob.},
month = dec,
pages = {4297–4311},
numpages = {15}
}

@inproceedings{nemati2016stability,
author = {Nemati, Alireza and Kumar, Rumit and Kumar, Manish},
year = {2016},
month = {10},
pages = {V001T05A005},
title = {Stabilizing and Control of Tilting-Rotor Quadcopter in Case of a Propeller Failure},
doi = {10.1115/DSCC2016-9897}
}

@INPROCEEDINGS{mallavalli2019fault,
  author={Mallavalli, Seema and Fekih, Afef},
  booktitle={2019 American Control Conference (ACC)}, 
  title={A Fault Tolerant Control Design for Actuator Fault Mitigation in Quadrotor UAVs}, 
  year={2019},
  volume={},
  number={},
  pages={5111-5116},
  doi={10.23919/ACC.2019.8815190}}

@article{mazare2024robust,
author = {Mazare, Mahmood and Taghizadeh, M. and Ghaf-Ghanbari, Pegah and Davoodi, Ehsan},
year = {2024},
month = {06},
pages = {104747},
title = {Robust Fault Detection and Adaptive Fixed-time Fault-Tolerant Control for Quadrotor UAVs},
volume = {179},
journal = {Robotics and Autonomous Systems},
doi = {10.1016/j.robot.2024.104747}
}

@article{abbaspour2020neural,
author = {Abbaspour, Alireza and Yen, Kang and Forouzannezhad, Parisa and Sargolzaei, Arman},
year = {2018},
month = {07},
pages = {1-11},
title = {A Neural Adaptive Approach for Active Fault-Tolerant Control Design in UAV},
volume = {PP},
journal = {IEEE Transactions on Systems, Man, and Cybernetics: Systems},
doi = {10.1109/TSMC.2018.2850701}
}

@article{yu2024fault,
author = {Yu, Hai and Wu, Shizhen and He, Wei and Liang, Xiao and Han, Jianda and Fang, Yongchun},
year = {2024},
month = {10},
pages = {12718-12731},
title = {Fault-Tolerant Control for Multirotor Aerial Transportation Systems With Blade Damage},
volume = {71},
journal = {IEEE Transactions on Industrial Electronics},
doi = {10.1109/TIE.2023.3347848}
}

@article{liang2024high,
author = {Liang, Weisheng and Chen, Zheng and Yao, Bin},
year = {2024},
month = {01},
pages = {1-14},
title = {High-Accuracy Adaptive Robust Fault-Tolerant Control for Quadrotor With Actuator Uncertainties and Aerodynamic Drag Compensation},
volume = {PP},
journal = {IEEE Transactions on Automation Science and Engineering},
doi = {10.1109/TASE.2024.3479294}
}

@article{miao2024fixed,
author = {Miao, Qiyang and Zhang, Ke and Jiang, Bin},
year = {2024},
month = {01},
pages = {},
title = {Fixed-Time Collision-Free Fault-Tolerant Formation Control of Multi-UAVs Under Actuator Faults},
volume = {PP},
journal = {IEEE transactions on cybernetics},
doi = {10.1109/TCYB.2024.3352251}
}

@article{wang2024event,
author = {Wang, Changhui and Li, Wencheng and Liang, Mei},
year = {2024},
month = {01},
pages = {1-18},
title = {Event-Triggered Prescribed Performance Adaptive Fuzzy Fault-Tolerant Control for Quadrotor UAV With Actuator Saturation and Failures},
volume = {PP},
journal = {IEEE Transactions on Aerospace and Electronic Systems},
doi = {10.1109/TAES.2024.3448404}
}

@article{hao2022fault,
  title={Fault-Tolerant Position Tracking Control Design for a Tilt Tri-Rotor Unmanned Aerial Vehicle},
  author={Wei Hao and Bin Xian and Tian Xie},
  journal={IEEE Transactions on Industrial Electronics},
  year={2022},
  volume={69},
  pages={604-612},
  url={https://api.semanticscholar.org/CorpusID:234203563}
}

@inproceedings{sohege2021novel,
author = {Sohege, Yves and Quinones-Grueiro, Marcos and Provan, Gregory},
year = {2021},
month = {05},
pages = {10719-10725},
title = {A Novel Hybrid Approach for Fault-Tolerant Control of UAVs based on Robust Reinforcement Learning},
doi = {10.1109/ICRA48506.2021.9562097}
}

@inproceedings{pourpanah2018anomaly,
  title={Anomaly detection and condition monitoring of UAV motors and propellers},
  author={Pourpanah, Farhad and Zhang, Bin and Ma, Rui and Hao, Qi},
  booktitle={2018 IEEE SENSORS},
  pages={1--4},
  year={2018},
  organization={IEEE}
}

@inproceedings{palanisamy2022fault,
  title={Fault detection and performance monitoring of propellers in electric UAV},
  author={Palanisamy, Rajendra P and Kulkarni, Chetan S and Corbetta, Matteo and Banerjee, Portia},
  booktitle={2022 IEEE Aerospace Conference (AERO)},
  pages={1--6},
  year={2022},
  organization={IEEE}
}

@inproceedings{ghalamchi2018vibration,
  title={Vibration-based propeller fault diagnosis for multicopters},
  author={Ghalamchi, Behnam and Mueller, Mark},
  booktitle={2018 International Conference on Unmanned Aircraft Systems (ICUAS)},
  pages={1041--1047},
  year={2018},
  organization={IEEE}
}

@inproceedings{mandralis2023minimum,
  title={Minimum time trajectory generation for bounding flight: Combining posture control and thrust vectoring},
  author={Mandralis, Ioannis and Sihite, Eric and Ramezani, Alireza and Gharib, Morteza},
  booktitle={2023 European Control Conference (ECC)},
  pages={1--7},
  year={2023},
  organization={IEEE}
}

@inproceedings{sihite2024dynamic,
  title={Dynamic modeling of wing-assisted inclined running with a morphing multi-modal robot},
  author={Sihite, Eric and Ramezani, Alireza and Gharib, Morteza},
  booktitle={2024 IEEE International Conference on Robotics and Automation (ICRA)},
  pages={2339--2345},
  year={2024},
  organization={IEEE}
}

@inproceedings{salagame2024quadrupedal,
  title={Quadrupedal Locomotion Control On Inclined Surfaces Using Collocation Method},
  author={Salagame, Adarsh and Gianello, Maria and Wang, Chenghao and Venkatesh, Kaushik and Pitroda, Shreyansh and Rajput, Rohit and Sihite, Eric and Leeser, Miriam and Ramezani, Alireza},
  booktitle={2024 American Control Conference (ACC)},
  pages={2838--2843},
  year={2024},
  organization={IEEE}
}

@inproceedings{krishnamurthy2024narrow,
  title={Narrow-path, dynamic walking using integrated posture manipulation and thrust vectoring},
  author={Krishnamurthy, Kaushik Venkatesh and Wang, Chenghao and Pitroda, Shreyansh and Salagame, Adarsh and Sihite, Eric and Nemovi, Reza and Ramezani, Alireza and Gharib, Morteza},
  booktitle={2024 IEEE International Conference on Advanced Intelligent Mechatronics (AIM)},
  pages={898--903},
  year={2024},
  organization={IEEE}
}

@inproceedings{sihite2024posture,
  title={Posture manipulation of thruster-enhanced bipedal robot performing dynamic wall-jumping using model predictive control},
  author={Sihite, Eric and Pitroda, Shreyansh and Liu, Taoran and Wang, Chenghao and Krishnamurthy, Kaushik Venkatesh and Salagame, Adarsh and Nemovi, Reza and Ramezani, Alireza and Gharib, Morteza},
  booktitle={2024 IEEE-RAS 23rd International Conference on Humanoid Robots (Humanoids)},
  pages={491--496},
  year={2024},
  organization={IEEE}
}
\end{document}